\documentclass[runningheads]{llncs}

\usepackage{appendix}
\usepackage{cite, makecell}
\usepackage{amsmath,amssymb,amsfonts}
\usepackage{algorithmic}
\usepackage{graphicx}
\graphicspath{{images/}}
\usepackage{textcomp}
\usepackage{xcolor}
\usepackage{balance}
\usepackage{multirow}
\usepackage{hyphenat}
\usepackage{multicol}

\def\BibTeX{{\rm B\kern-.05em{\sc i\kern-.025em b}\kern-.08em
		T\kern-.1667em\lower.7ex\hbox{E}\kern-.125emX}}
\usepackage{amssymb,amsmath,epsfig,latexsym, bm}
\usepackage{nohyperref}	
\hypersetup{colorlinks, pdfa=true, breaklinks, citecolor=black, urlcolor=black, linkcolor=black}
\usepackage{amssymb,amsmath,epsfig,latexsym, bm}
\usepackage{mathtools, mdwmath, mdwtab, multirow, amssymb, amsmath, array, ragged2e}

\usepackage{threeparttable, booktabs, url, eqparbox, xspace, setspace, tabularx, subfig}
\makeatletter
\DeclareRobustCommand\onedot{\futurelet\@let@token\@onedot}
\def\@onedot{\ifx\@let@token.\else.\null\fi\xspace}

\def\etal{\emph{et al}\onedot}
\makeatother

\begin{document}
\title{Performance Analysis of Various EfficientNet Based U-Net++ Architecture for Automatic Building Extraction from High Resolution Satellite Images}
\author{\vspace{.85in}}


 \author{Tareque Bashar Ovi\orcidID{0000-0001-6961-8894} \and Nomaiya Bashree\orcidID{0009-0008-6151-2356} \and Protik Mukherjee\orcidID{0009-0007-9710-8868} \and Shakil Mosharrof\orcidID{0000-0002-7228-2823} \and Masuma Anjum Parthima\orcidID{0009-0007-3269-144X}}

\authorrunning{T. B. Ovi~\etal}
\titlerunning{Satellite Image Segmentation}

%

 \institute{Military Institute of Science and Technology (MIST)\\
 Mirpur Cantonment, Dhaka--1216, Bangladesh\\
 \email{ovitareque@gmail.com}\hfill
 \email{nomaiyabashree2002@gmail.com}\hfill
 \email{protik.eece@gmail.com}\hfill
 \email{shakilmrf8@gmail.com}\hspace{0.5cm}
 \email{masumaanjum48@gmail.com}
 }

\maketitle              
\begin{abstract}
Building extraction is an essential component of study in the science of remote sensing, and applications for building extraction heavily rely on semantic segmentation of high-resolution remote sensing imagery. Semantic information extraction gap constraints in the present deep learning based approaches, however can result in inadequate segmentation outcomes. To address this issue and  extract buildings with high accuracy, various efficientNet backbone based U-Net++ has been proposed in this study. The designed network, based on U-Net, can improve the sensitivity of the model by deep supervision, voluminous  redesigned skip-connections and hence reducing the influence of irrelevant feature areas in the background. Various effecientNet backbone based encoders have been employed when training the network to enhance the capacity of the model to extract more relevant feature. According on the experimental findings, the suggested model significantly outperforms previous cutting-edge approaches. Among  the $5$ efficientNet variation Unet++ based on efficientb4 achieved the best result by scoring mean accuracy of $92.23$\%, mean iou of $88.32$\%, and mean precision of $93.2$\% on publicly available Massachusetts building dataset and thus showing  the promises of the model for automatic building extraction from high resolution satellite images.

\end{abstract}


\section{Introduction}
Estimating population density, urban planning, and the creation and updating of topographic maps all rely on the automatic recognition and building extraction from remote-sensing photos. Despite the attention that building extraction has gotten, it is still a difficult operation because of the noise, occlusion, and intricacy of the background in the original remote sensing images.  Buildings can be extracted from remote sensing images using a number of different techniques that have been developed recently.Deep convolutional neural networks (CNN) advancements have led to a revolution in the automatic extraction of cartographic information from extremely high resolution aerial and satellite imagery. The fundamental benefit of these supervised CNNs is that they can automatically learn features from training inputs with little to no task-specific information.
 The accuracy of CNN is comparable to that of human classification accuracy, but it is constant and quick, allowing for quick application over very vast areas and/or over time.These CNNs could facilitate the quick collection of precise spatial data on city buildings and, in turn, the creation of building environment maps, which are crucial for urban planning and monitoring. Two forms of segmentation with CNN can be used for this building extraction task: instance segmentation and semantic segmentation. A class is assigned to each pixel in an image as part of semantic segmentation. When creating segmentation from extremely high resolution images, this kind of segmentation has an adequate degree of precision. A MultiRes-UNet network was suggested by Abolfazl Abdollahi\etal~\cite{abdollahi2021integrating} for the extraction of buildings from aerial photographs. Their performance was remarkable, with an F1 score of $96.98$\%, an MCC of $95.73$\%, and an IOU of $94.13$\% using the AIRS dataset. Using the WHU Building Dataset and Urban3d Challenge dataset, LEILEI XU\etal~\cite{xu2021ha} presented the Holistically-Nested Attention U-Net (HA U-Net) for building segmentation. On the Urban3d dataset, they obtained an IOU of $70.66$\% and a Kappa of $80.21$\%; on the WHU Building Dataset, they obtained an IOU of $72.74$\% and a Kappa of $79.42$\%. Waleed Alsabhan\etal~\cite{alsabhan2022automatic} used the Massachusetts building dataset to create a U-net architecture for semantic segmentation. Using Unet-ResNet50, they reported an IOU score of $82.2$\% and an accuracy of $90.2$\%, whereas using the traditional U-Net, they recorded an IOU score of $23.16$\% and an accuracy of $71.9$\%. FCN, Segnet, Deeplab V3, and ENRU approaches had been deployed. With these techniques, they attained respective OAs of $93.37$\%, $93.84$\%, $93.01$\%, and $94.12$\%, as well as IoUs of $69.47$\%, $72.1$\%, respectively $68.55$\%, and $72.77$\%. Using WorldView-2 satellite remote sensing picture datasets, Chuangnong Li\etal~\cite{li2021attention} suggested an attention-enhanced U-Net for building extraction from agriculture. With their model, they attained an accuracy of $96.96$\%, an F1 score of $81.47$\%, a recall of $82.72$\%, and an IOU of $68.72$\%.
 Ibrahim Delibasoglu\etal~\cite{delibasoglu2020improved} used the Massachusetts building dataset along with the Ikonos and Quickbird pan-sharpened satellite image collection to design an Inception UNet-v2 architecture for building detection. On the Ikonos dataset, they acquired a precision of $88.97$\% and an F1 score of $82.03$\%. On the Massachusetts building dataset, they attained a precision of $73.69$\% and an F1 score of $78.39$\%. Mehdi Khoshboresh-Masouleh\etal~\cite{khoshboresh2020multiscale} suggested a deep dilated CNN for developing extraction from the Indiana, WHU-I, Inriaa, and Potsdama datasets. They performed well, earning F1 scores between $80$\% and $96$\% and IOU scores between $67$\% and $92$\%. DeepResUnet was presented by Yaning Yi\etal~\cite{yi2019semantic} for the segmentation of urban buildings from aerial pictures. Using their model, they were able to attain accuracy of $94.01$\%, recall of $93.28$\%, an F1 score of $93.64$\%, and a Kappa of $91.76$\%.
By using WorldView-3 images for generating instance segmentation, Fabien H. Wagner\etal~\cite{wagner2020u} obtained an overall accuracy of $97.67$\% using the U-net architecture.

According to current literature, there have been various attempts to semantically segment buildings from satellite images using standard end-to-end deep learning models. However, a complete experiment based on several efficientNet backbones based U-Net++ has not yet been conducted. The usage of labelled data has once again shown that deep learning-based systems have significantly improved. They don't perform well enough with unlabeled data to qualify as state-of-the-art. Therefore our contribution in the paper can be reported in the way specified below after taking everything into account:
\renewcommand{\labelenumi}{\alph{enumi}}
\begin{enumerate}
    \item Proposing an end-to-end deep learning based solution for automatic extraction of buildings from high resolution satellite images with high accuracy using U-Net++ architecture.
    \item Evaluating a comparative performance analysis among various efficientNet backbones as the encoder for U-Net++ architecture to maximize the performance.
\end{enumerate}

\section{Dataset}
\subsection{Dataset Description}
The Massachusetts Buildings Dataset~\cite{MnihThesis} is used in this study which includes $151$ aerial images, each having $1500\times1500$ pixels resolution. The dataset is divided into three sets: a training set of $137$ images, a test set of $10$ images, and a validation set of $4$ images.
\begin{figure}
\label{fig:dataset}
\centering
	
		\subfloat[]{
		\label{fig:im1}
		\includegraphics[width=0.21\columnwidth]{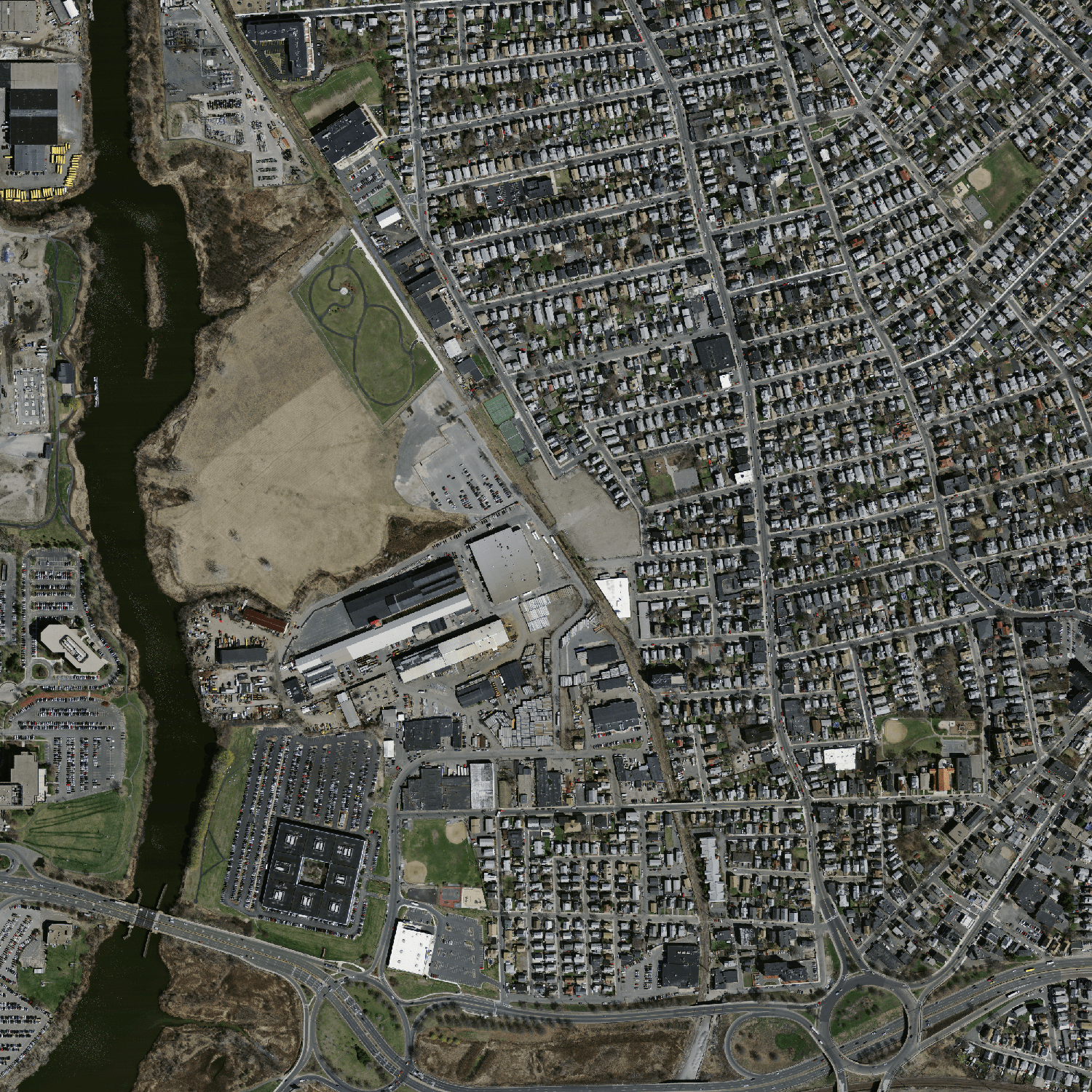}
	}	
	\subfloat[]{
		\label{fig:im2}
		\includegraphics[width=0.21\columnwidth]{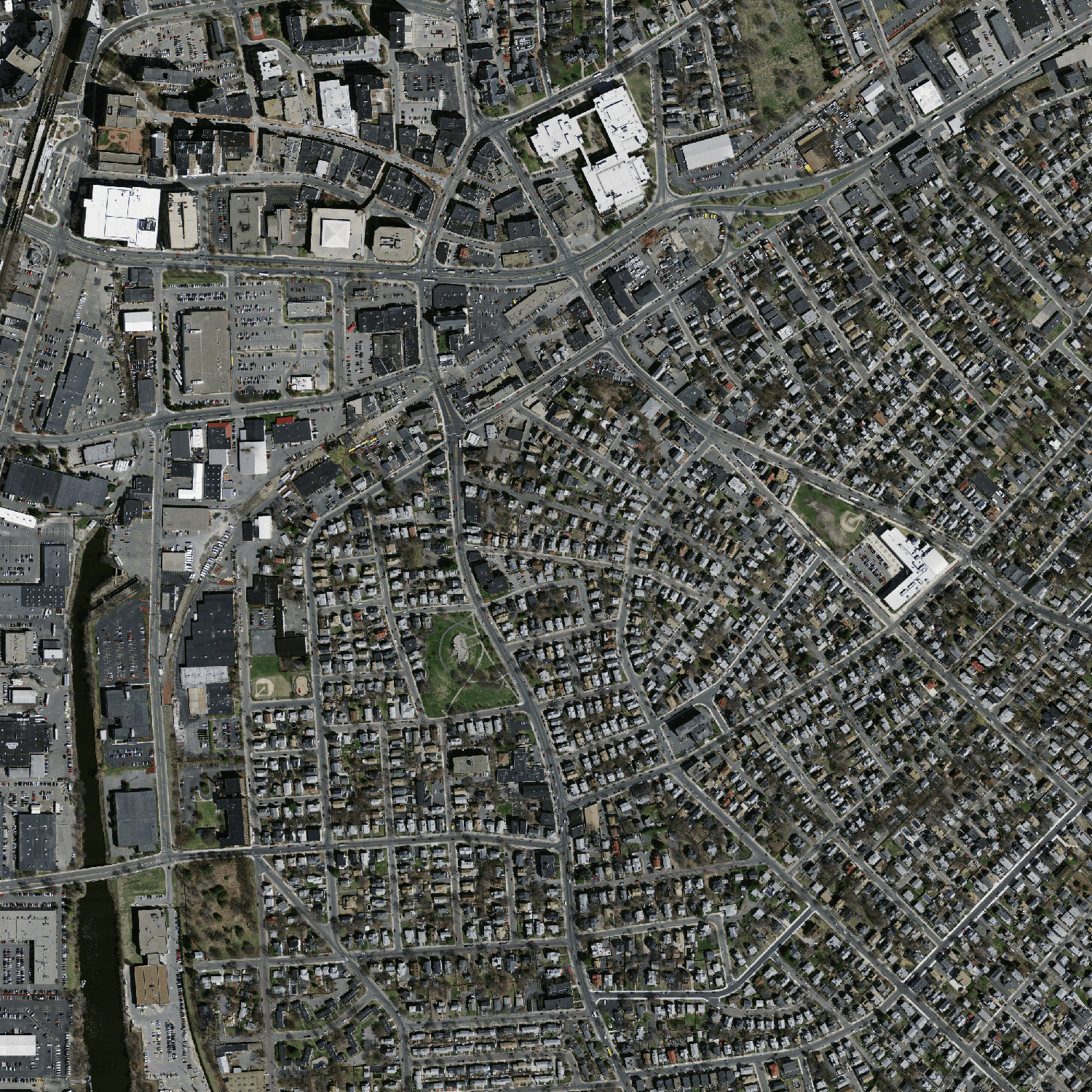}
	
	}	
	\subfloat[]{
		\includegraphics[width=0.21\columnwidth]{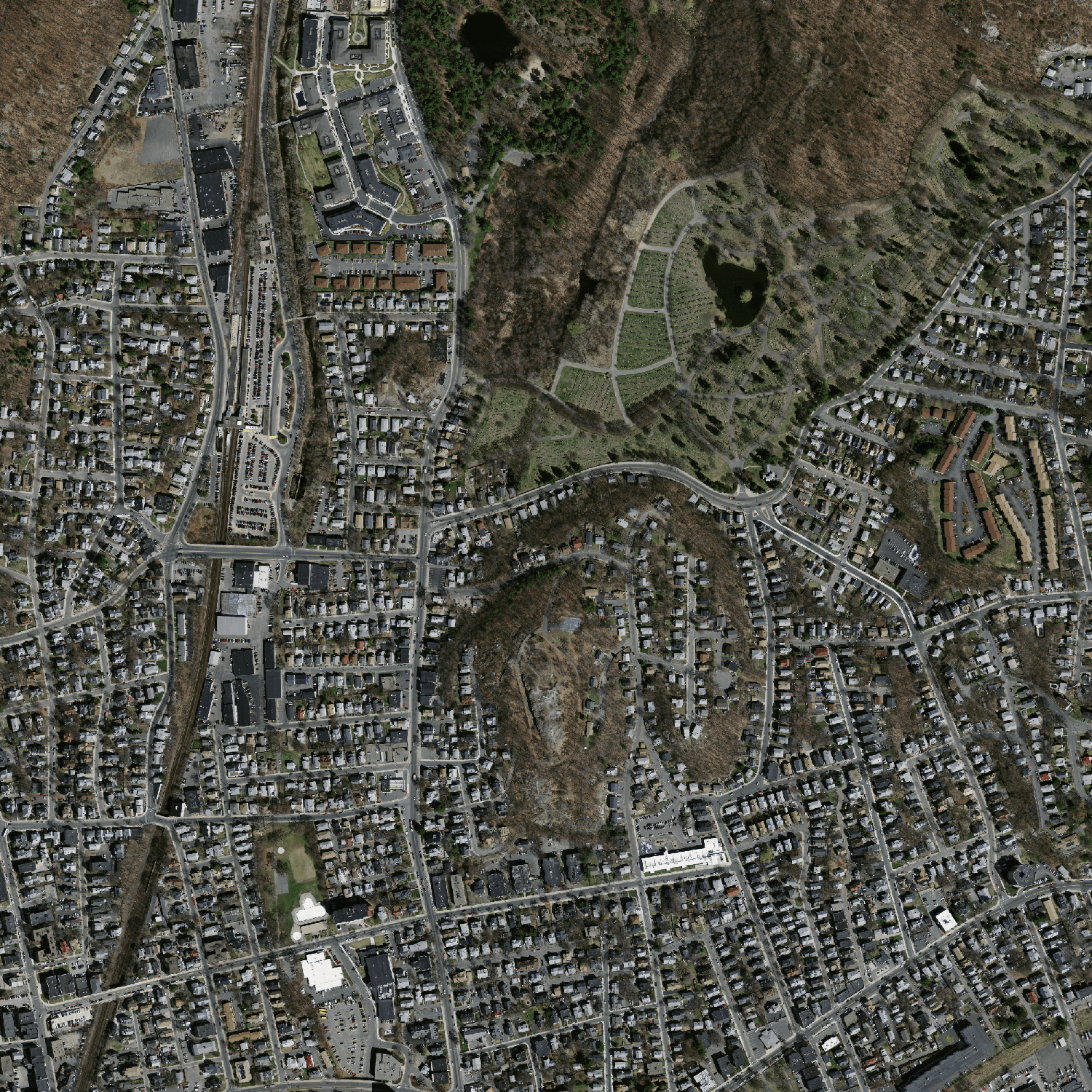}
		\label{fig:im3}
	}\\
		\subfloat[]{
		\includegraphics[width=0.21\columnwidth]{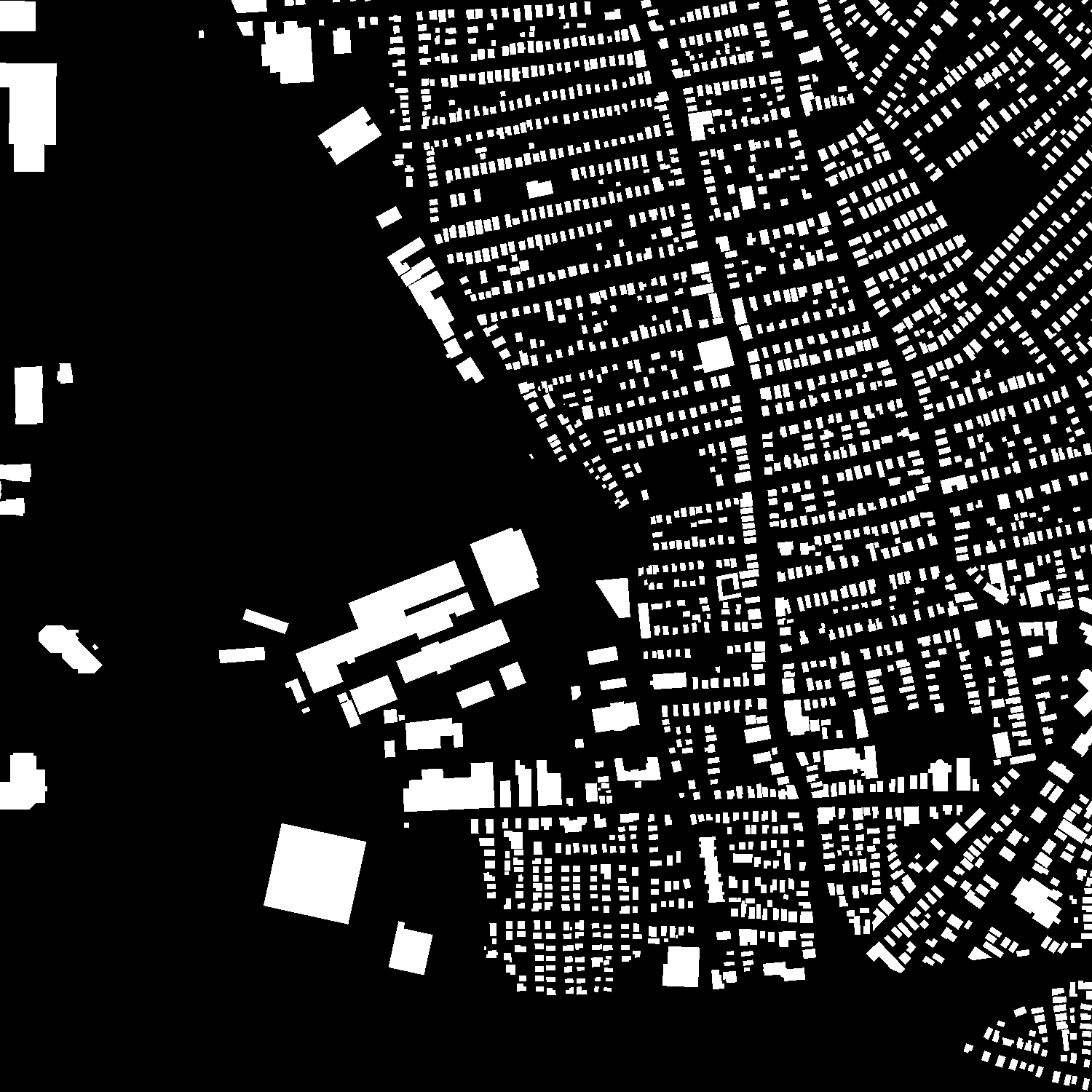}
		\label{fig:gt1}
	}
		\subfloat[]{
		\includegraphics[width=0.21\columnwidth]{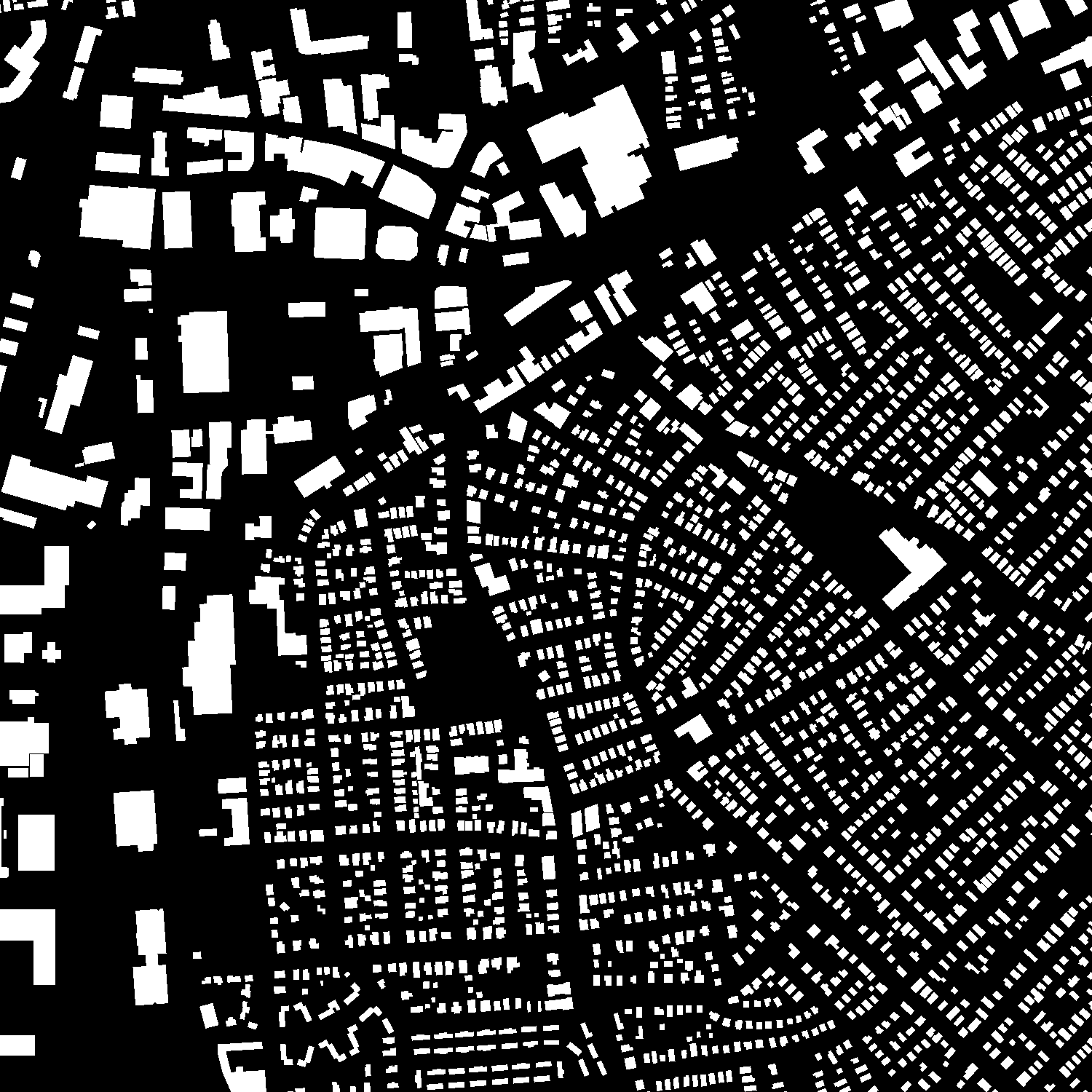}
		\label{fig:gt2}
	}
	\subfloat[]{
		\label{fig:gt3}
		\includegraphics[width=0.21\columnwidth]{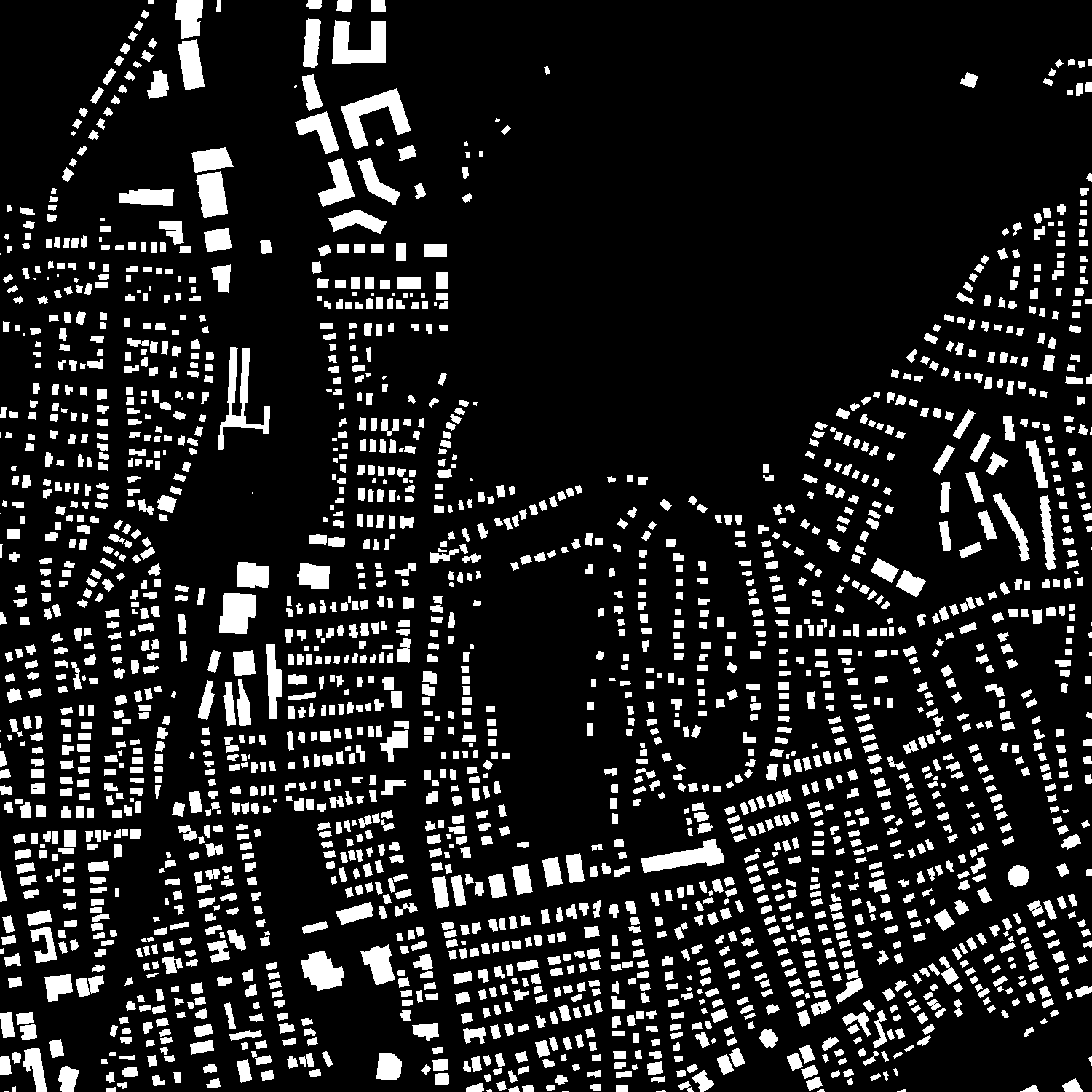}
	} 	
	
	\caption{Example of images from the dataset: (a--c)~Input image and (d--f)~Target image}

\end{figure}

\subsection{Dataset Preprocessing}
At first all the images and the ground truths are  down-sampled to $256\times256\times3$ and the only  preprocessing step is to normalize the pixel values from $-1$ to $1$ as the Equation \eqref{norm}. Here $i, j$ represents image height and width respectively, $I_{i,j}$ is the original image and $I^{n}_{i,j}$ is the normalized image. No other prior preprocessing step is required for this study.

\begin{equation}
    \label{norm}
    I^{n}_{i,j} = \frac{I_{i,j}}{127.5}-1
\end{equation}

At first both the training images and ground truths are down-sampled to $256\times256\times3$ and normalized. After that these images are used for the network training. When the whole training is done the best weight was saved for test purpose and evauluation.

\section{Methodology}
\begin{figure*}[htbp]
    \centering
    \includegraphics[width=1.0\columnwidth]{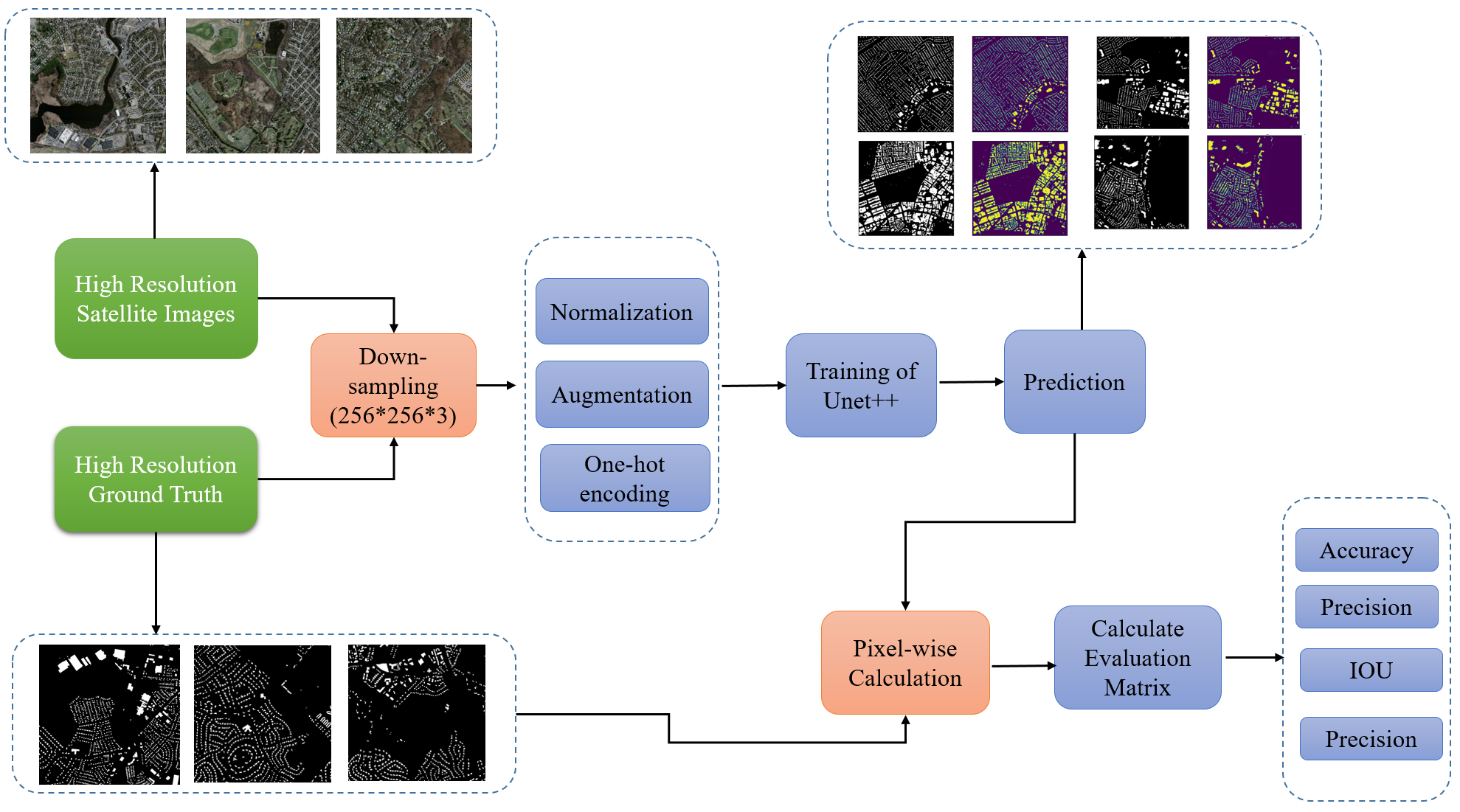}
    \caption{Research and Evaluation Worlflow}
    \label{work-flow}
\end{figure*}

The following steps illustrate our study and evaluation process.
\renewcommand{\labelenumi}{\alph{enumi}}
\begin{enumerate}
    \item At first both the training images and ground truths were down-sampled to $256\times256\times3$ from the high resolution satellite images.
    \item After that these images were normalized and then necessary augmentation and one hot encoding were performed for the training of the proposed network.
    \item After the training is complete, the best stored weight generates high resolution predictions, which are then evaluated using pixel-by-pixel calculations. In Fig.\ref{work-flow}, the entire process of this study is depicted.
\end{enumerate}

Below, for better understanding of the proposed methodology, is a brief explanation of U-Net++,EfficientNet backbone architecture.

\subsubsection{U-Net++} Using the U-Net as its foundation, UNet++ proposed by Zongwei Zhou\etal~\cite{zhou2018unet}, is an architecture for semantic segmentation. It improves the extraction of features by utilizing densely linked nested decoder sub-networks. Dense block and convolution layers are being added by UNet++ between the encoder and decoder to further enhance segmentation precision, which is crucial for medical imaging because even small segmentation errors could result in inaccurate results, which would be marginalized in clinical settings. The newly developed skip connections have been introduced by UNet++ in order to bridge the semantic gap between the encoder and decoder subpaths.These convolutional layers are designed to fill the semantic gaps between the encoder and decoder sub-networks' feature maps. As a result, the optimiser may be faced with a simpler optimization task. Moreover we can summarize the core features of U-Net++ as follows:

\begin{enumerate}
    \item redesign of the skip connections.
    \item voluminous skip connections.
    \item extensive supervision.
\end{enumerate}
\begin{figure*}[htbp]
    \centering
    \includegraphics[width=1.0\columnwidth]{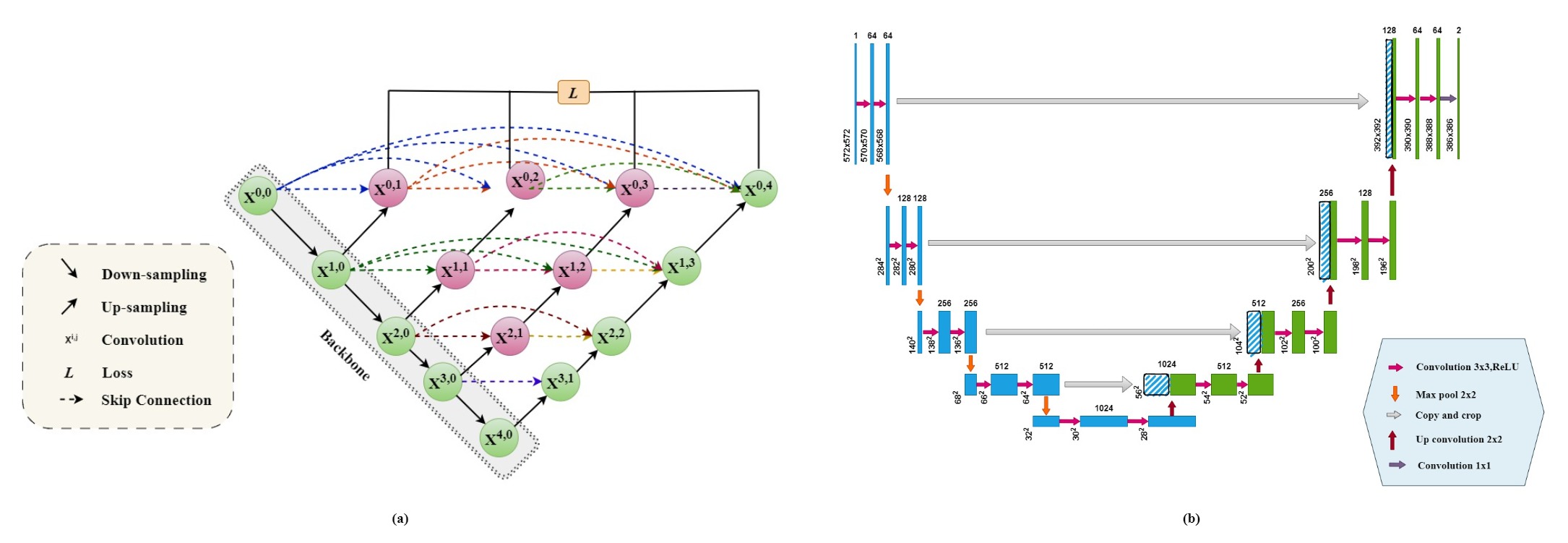}
    \caption{U-Net++ Vs Unet Architecture}
    \label{U-Net++_architecture}
\end{figure*}

\begin{figure}[htbp]
    \centering
    \includegraphics[width=0.6\columnwidth]{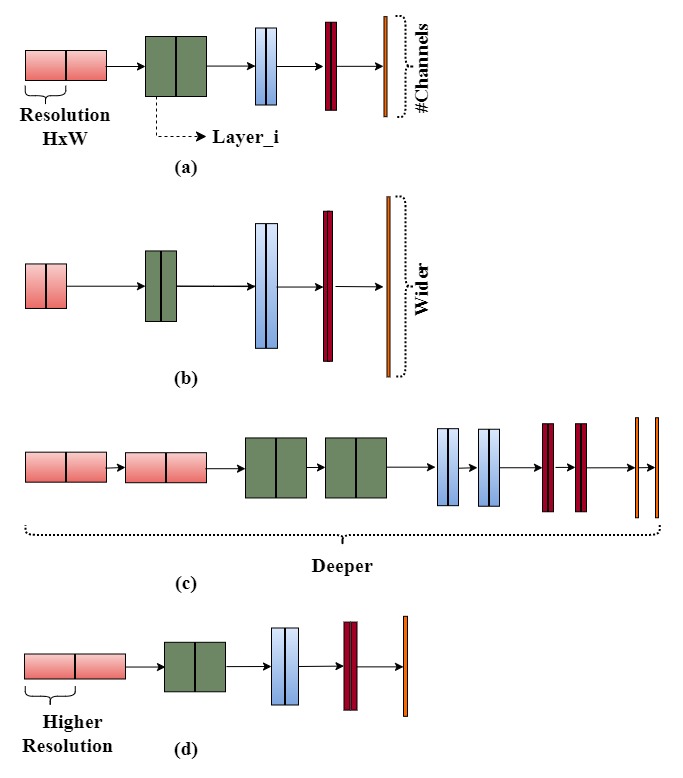}
    \caption{EfficientNet(Model Architecture)}
    \label{efficient_architecture}
\end{figure}
\subsubsection{Proposed Architecture}In this study, performance of total $5$ efficientNet backbone based Unet++ architecture has been analysed  for building density segmentation from high resolution satellite image with high accuracy. Unet++ has been considered for the segmentation architecture instead of Unet because,
\renewcommand{\labelenumi}{\alph{enumi}}
\begin{enumerate}
    \item In order to merge the semantic feature gap between contracting and decoder feature maps, convolution layers on skip routes are being used.
    \item Numerous skip connections based skip pathways are designed to enhance the gradient flow.
    \item having extensive supervision, which allows for model pruning and improves performance or, in the worst scenario, obtains performance comparable to utilizing just one loss layer.
    \item By merging otherwise semantically differing feature maps, the U-Net skip connections connect the feature maps directly between the encoder and the decoder.
    \item UNet++, on the other hand, combines the output of the preceding convolution layer of the same dense block with the matching up-sampled output of the lower dense block. The semantic level of the encoded feature is raised to be closer to that of the feature maps waiting in the decoder when receiving feature maps with equivalent semantic qualities, making optimization easier.
\end{enumerate}
EfficientNet has been chosen for the feature extraction for the following reason,
\renewcommand{\labelenumi}{\alph{enumi}}
\begin{enumerate}
    \item For feature extraction, the compound scaling approach significantly increased the model's accuracy and efficiency compared to earlier CNN models like MobileNet and ResNet..
    \item the since efficientNet models are designed by neural architecture search, utilizing them as encoders was significantly more computationally effective.

\end{enumerate}

\section{Result Analysis}
\begin{figure}
\begin{multicols}{2}
\label{fig:dataset}
	
		\subfloat[]{
		\label{fig:im1}
		\includegraphics[width=1.17\columnwidth]{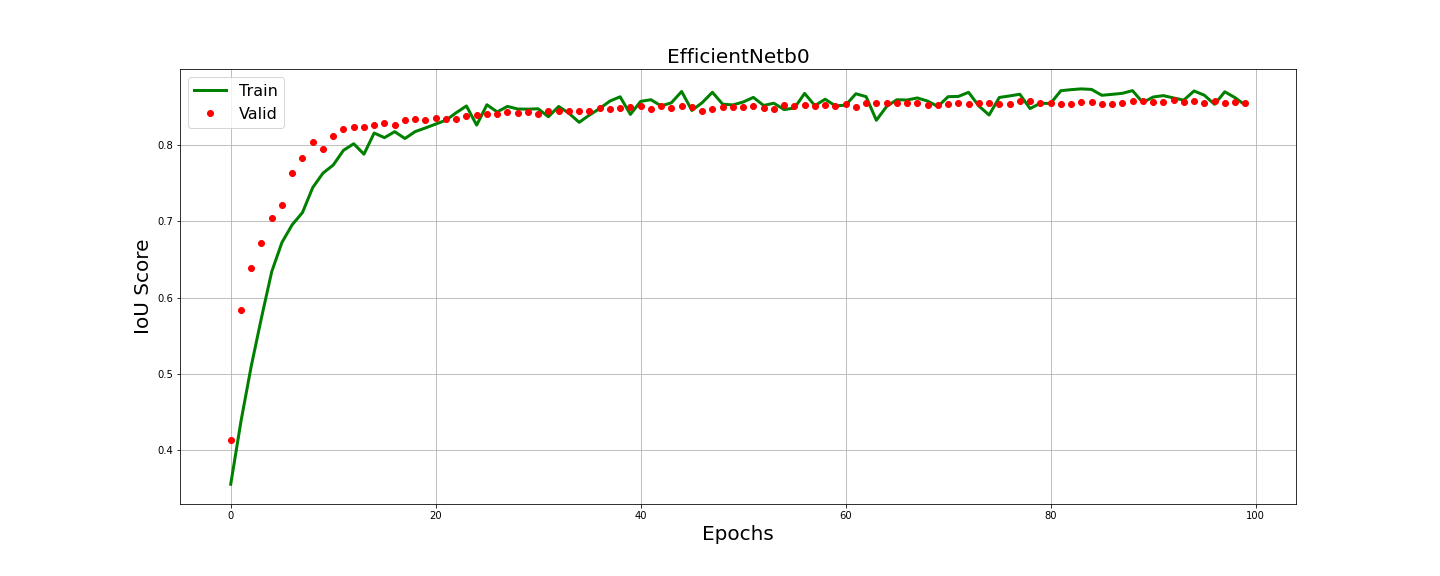}}
		
	\subfloat[]{
		\label{fig:im2}
		\includegraphics[width=1.16\columnwidth]{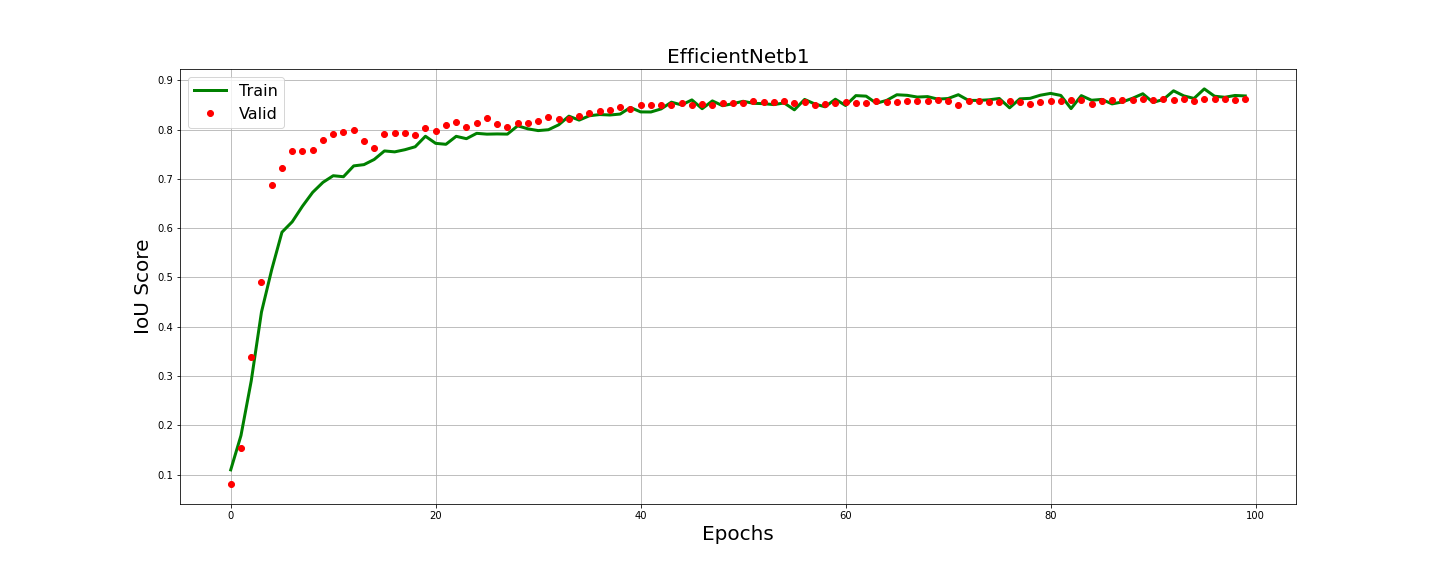}}

	\subfloat[]{
		\includegraphics[width=1.16\columnwidth]{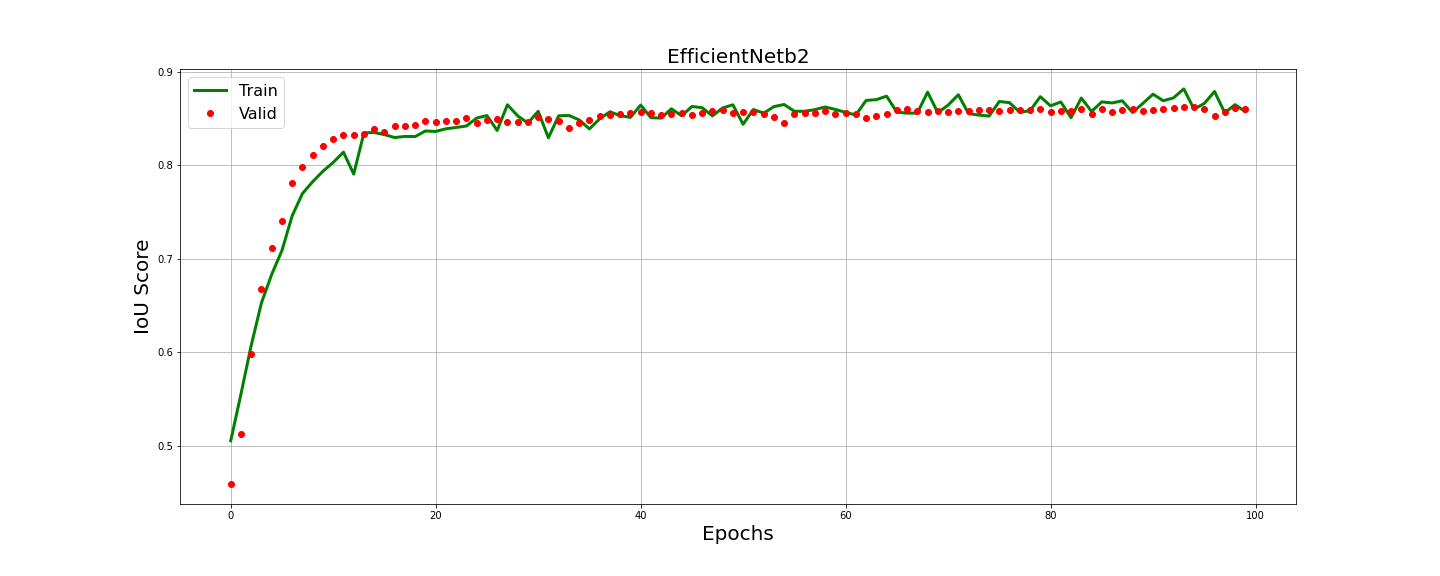}}
		\label{fig:im3}
	\\
		\subfloat[]{
		\includegraphics[width=1.16\columnwidth]{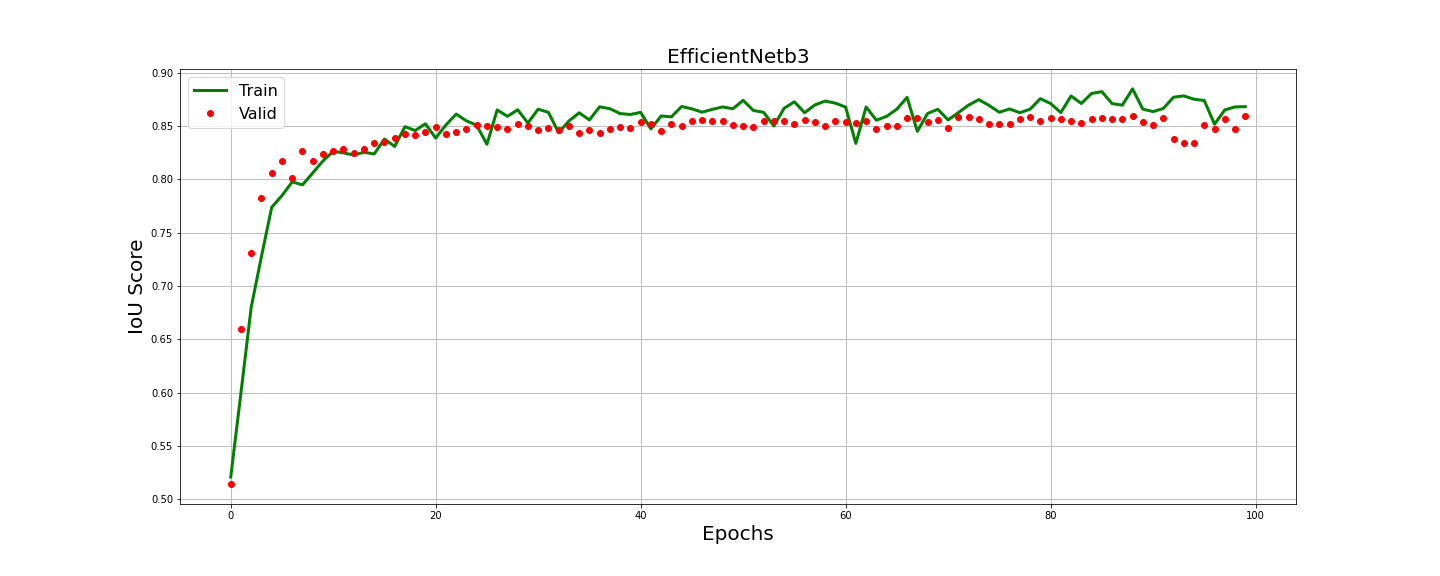}}
		\label{fig:gt1}
	
		\subfloat[]{
		\includegraphics[width=1.16\columnwidth]{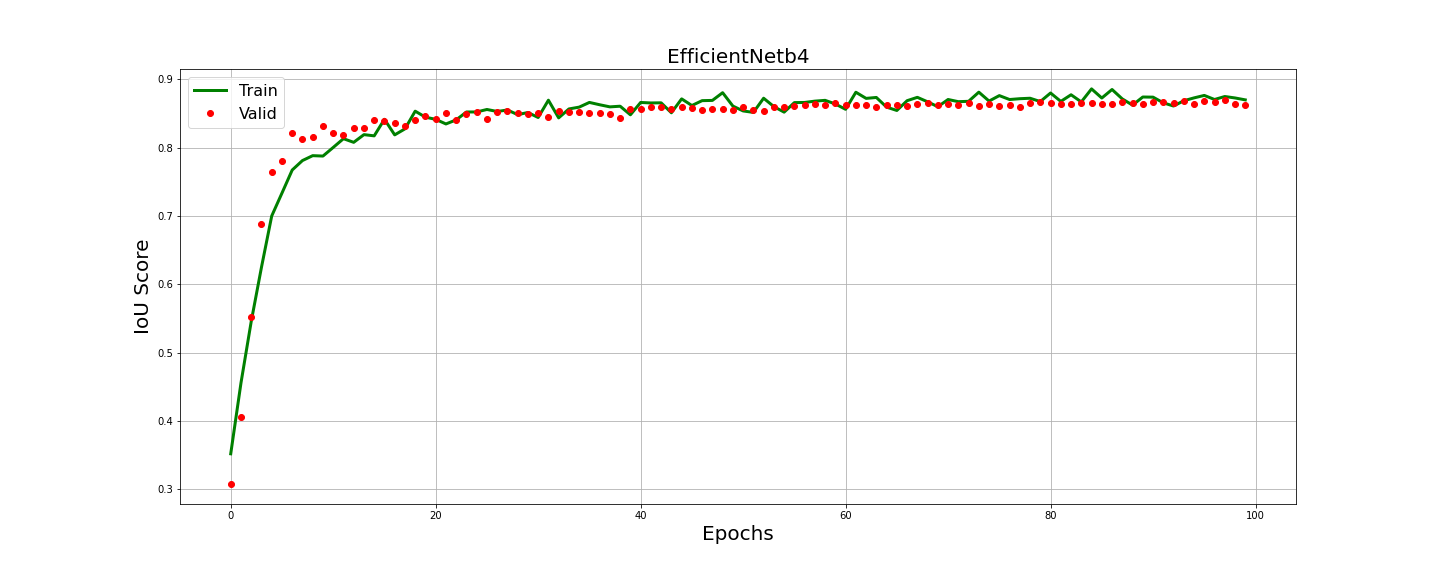}}
		\label{fig:gt2}
		
	\caption{Training History of IOU score}
        \label{iou_history}
		\subfloat[]{
		\label{fig:im1}
		\includegraphics[width=1.13\columnwidth]{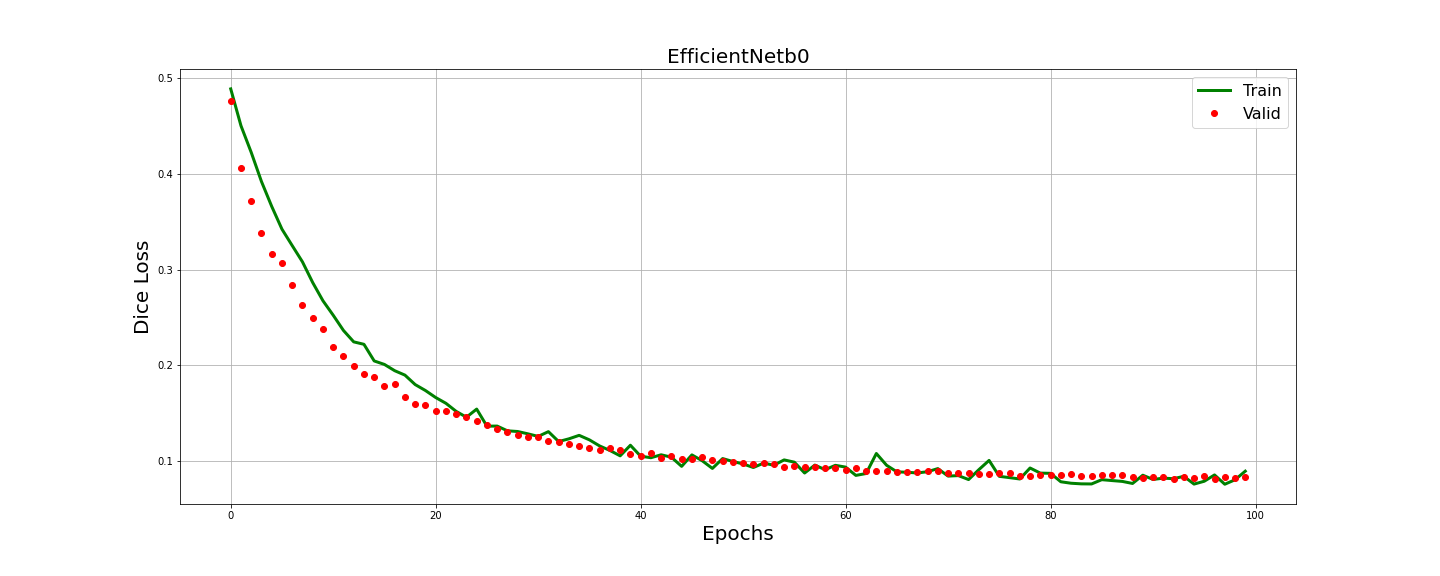}}
		
	\subfloat[]{
		\label{fig:im2}
		\includegraphics[width=1.13\columnwidth]{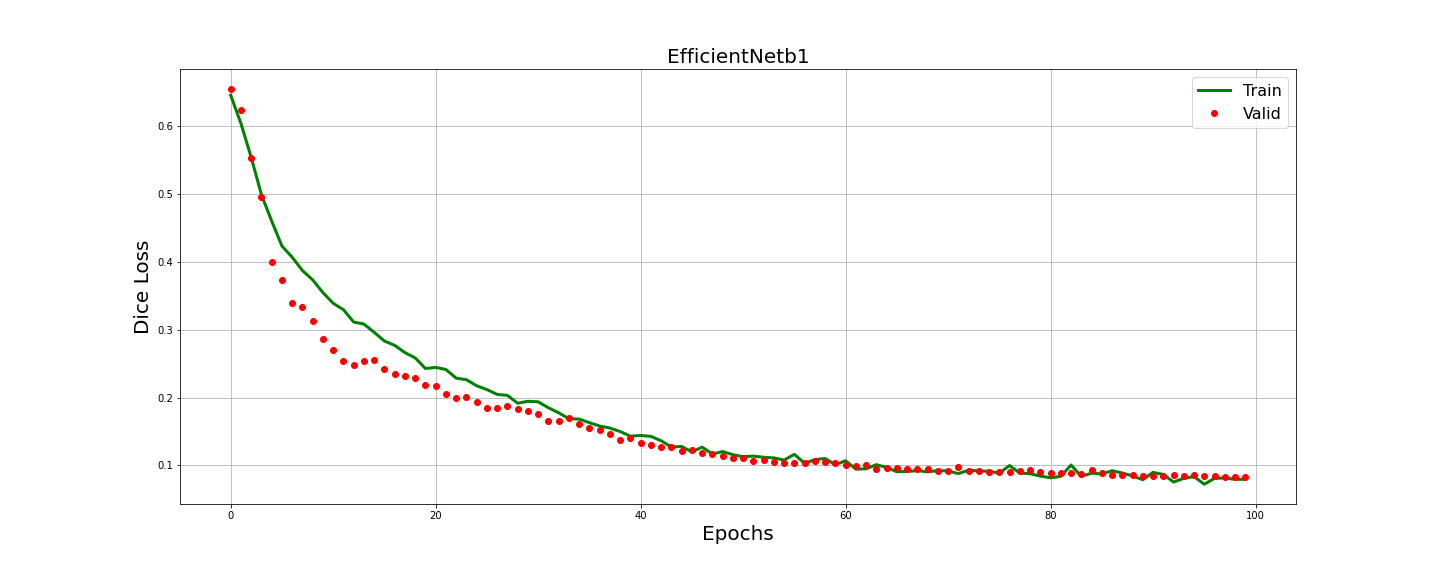}}

	\subfloat[]{
		\includegraphics[width=1.13\columnwidth]{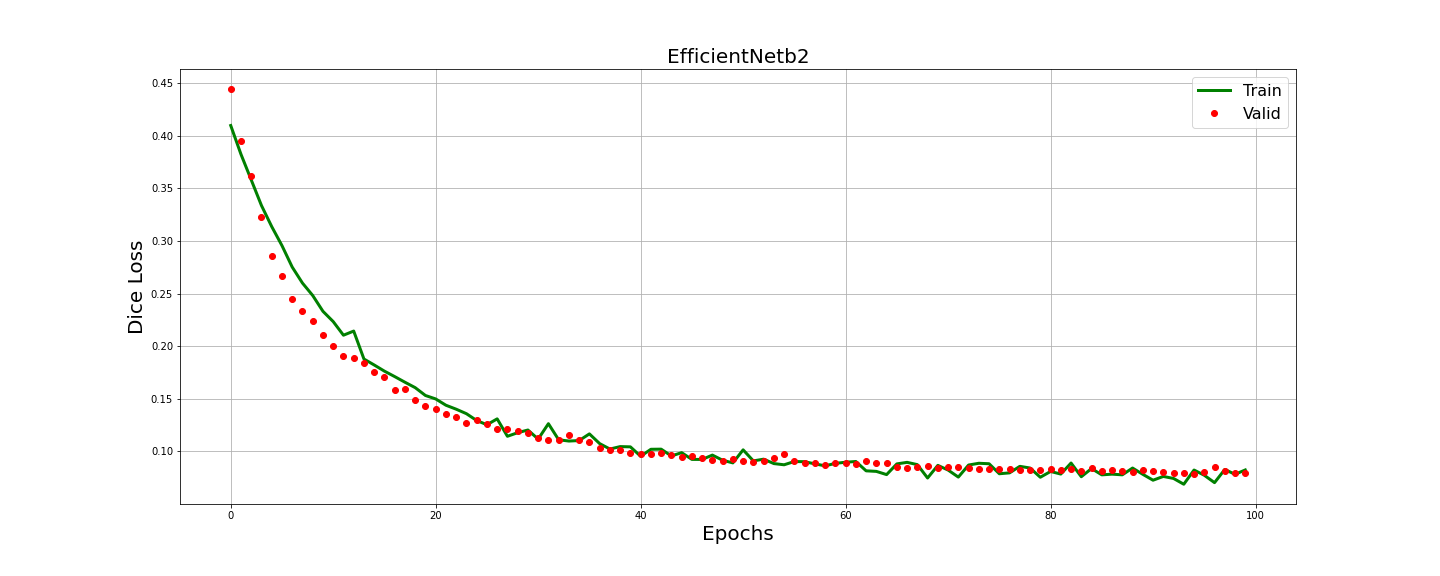}}
		\label{fig:im3}
	\\
		\subfloat[]{
		\includegraphics[width=1.13\columnwidth]{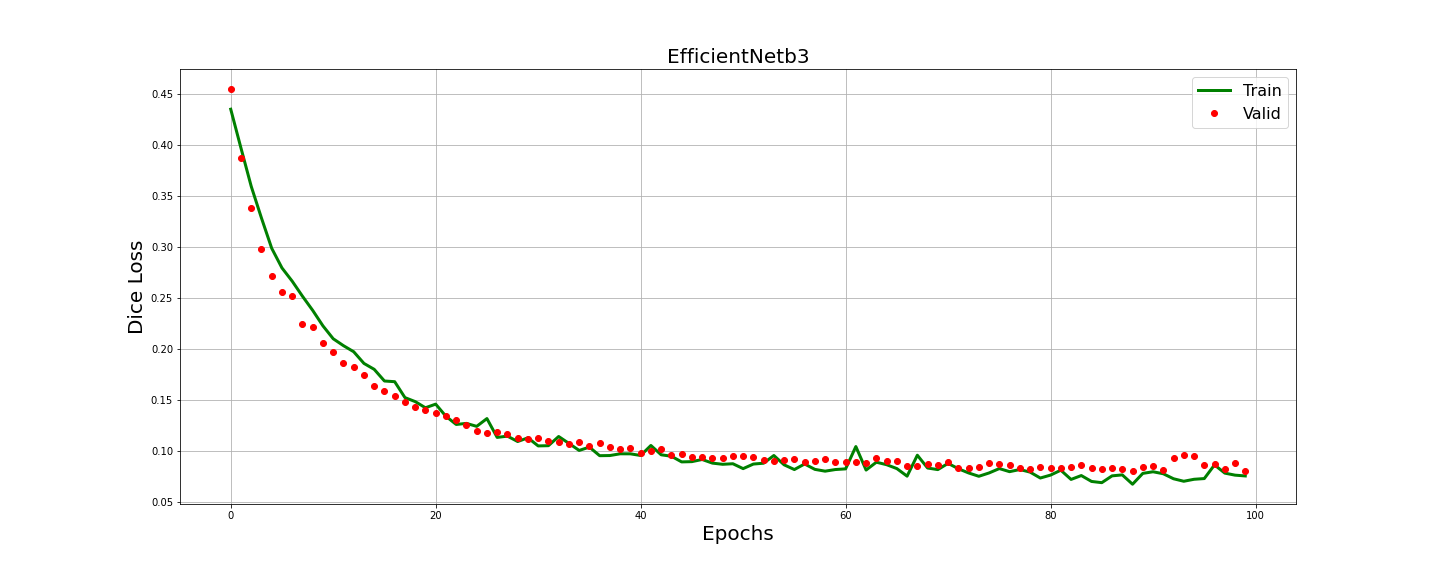}}
		\label{fig:gt1}
	
		\subfloat[]{
		\includegraphics[width=1.13\columnwidth]{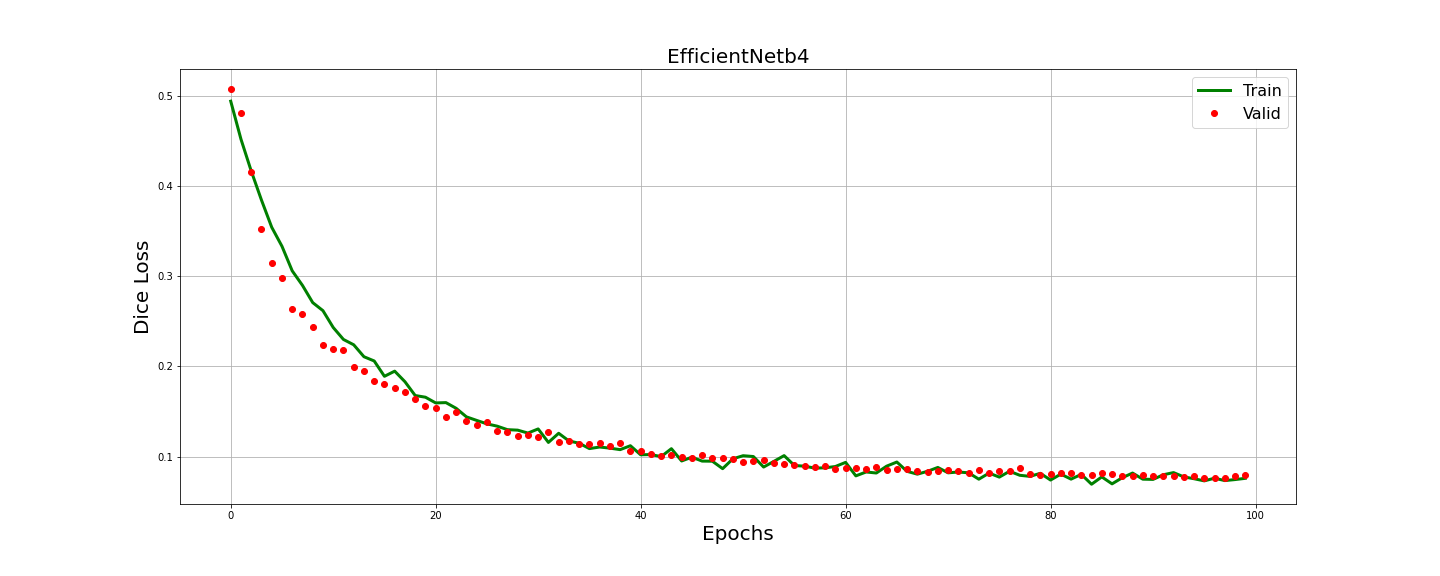}}
		\label{fig:gt2}
		
	\caption{Training History of Dice Loss}
        \label{loss_history}
\end{multicols}

\end{figure}

The performance of the models was evaluated based on accuracy, precision, recall, and IoU. Accuracy represents the percentage of correctly classified pixels, precision measures the model's ability to correctly identify building pixels, recall measures the proportion of actual building pixels correctly identified by the model, and IoU calculates the overlap between the predicted and ground truth masks. Fig.\ref{iou_history} and Fig.\ref{loss_history} illustrate the training history of IoU score and dice loss for the proposed Unet++ architecture with different EfficientNet-based encoders. The plots show the progress of the evaluation metrics during the training process. Notably, there was no overfitting observed, as the curves for all evaluation parameters demonstrated a steady improvement.

\begin{figure}
\label{fig:dataset}
	
		\subfloat[]{
		\label{fig:im1}
		\includegraphics[width=0.71\columnwidth]{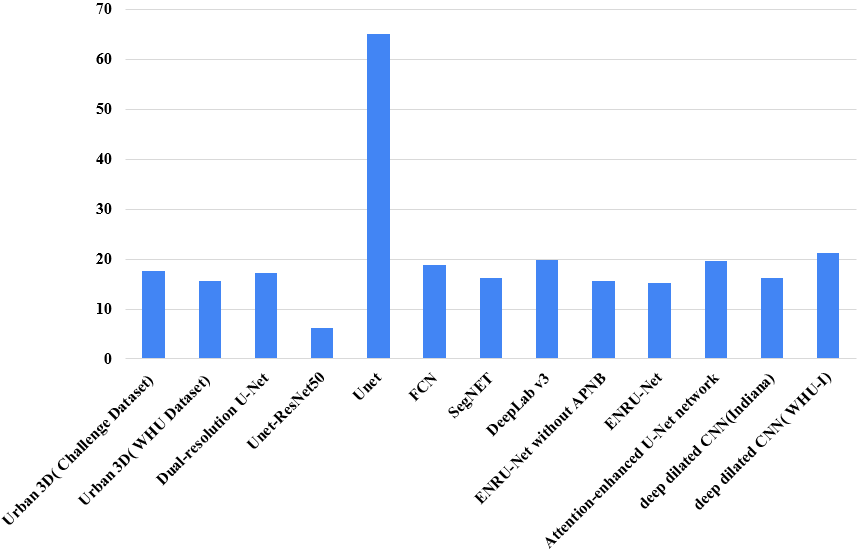}
	}	
	\subfloat[]{
		\label{fig:im2}
		\includegraphics[width=0.4\columnwidth]{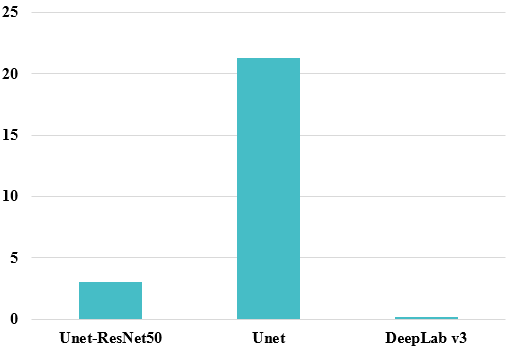}
	
	}\\	
	\subfloat[]{
		\includegraphics[width=0.55\columnwidth]{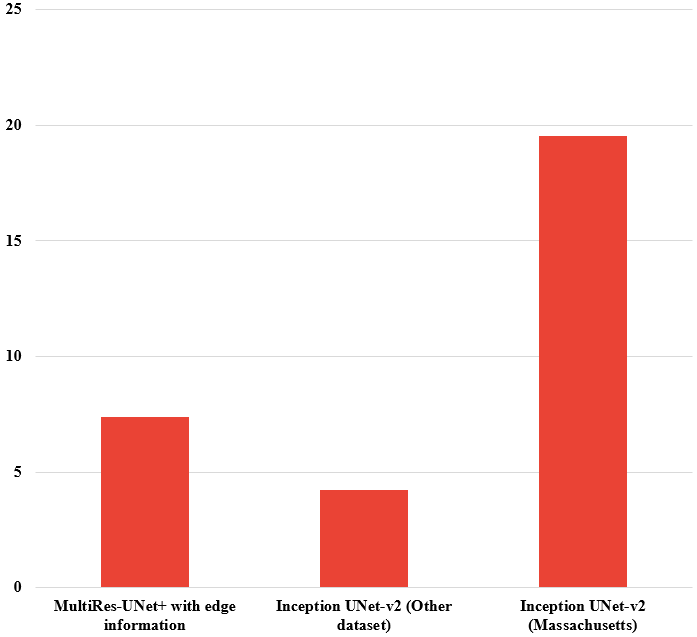}
		\label{fig:im3}
	}
		\subfloat[]{
		\includegraphics[width=0.5\columnwidth]{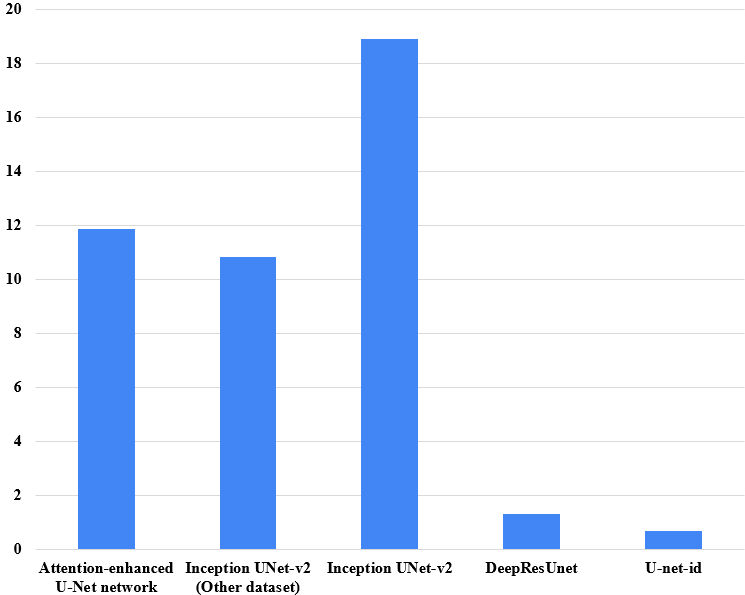}
		\label{fig:gt1}
	}

	\caption{Improvement of U-Net++(efficientNetb4) over other methods (a)~IOU (b)~Accuracy (c)~Precision and (d)~Recall }
\label{improvement}
\end{figure}

In Fig.\ref{improvement}, a improvement is shown among the proposed EfficientNet-based U-Net++ models with other popular architectures.
The proposed EfficientNet-based U-Net++ models consistently outperformed most of the compared models in terms of IoU, accuracy, recall and precision. Based on the experimental findings, the EfficientNetb4-based U-Net++ model attained the highest performance among all variants.

However, Table.\ref{performance_table} presents a performance comparison between the proposed EfficientNet-based U-Net++ models and some state-of-the-art approaches for building density segmentation. The proposed models achieved remarkable results, outperforming the existing literature across all evaluation parameters by a significant margin. The superior performance of the proposed models demonstrates the effectiveness of incorporating EfficientNet backbones and U-Net++ architecture for building extraction tasks.

\begin{figure}
\label{fig:dataset}
	
		\subfloat[]{
		\label{fig:im1}
		\includegraphics[width=1.0\columnwidth]{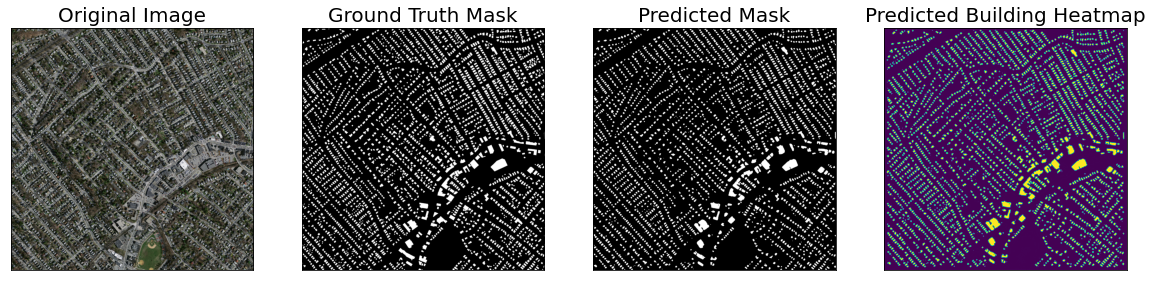}}
		
	\subfloat[]{
		\label{fig:im2}
		\includegraphics[width=1.0\columnwidth]{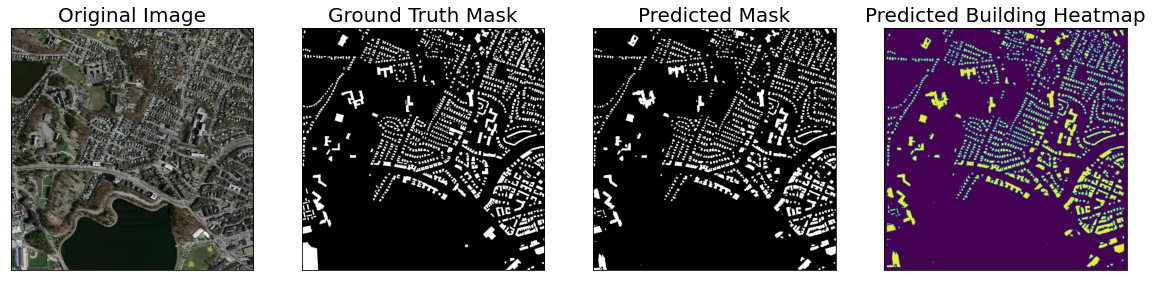}}

	\subfloat[]{
		\includegraphics[width=1.0\columnwidth]{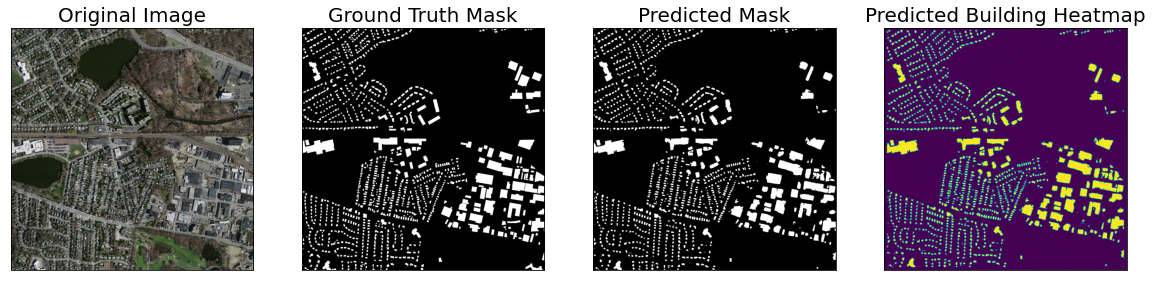}}
		\label{fig:im3}
	\\
		\subfloat[]{
		\includegraphics[width=1.0\columnwidth]{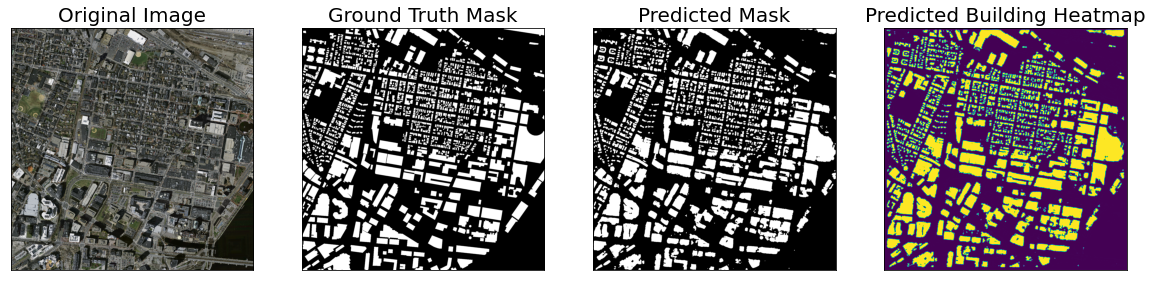}}
		\label{fig:gt1}
	
		\subfloat[]{
		\includegraphics[width=1.0\columnwidth]{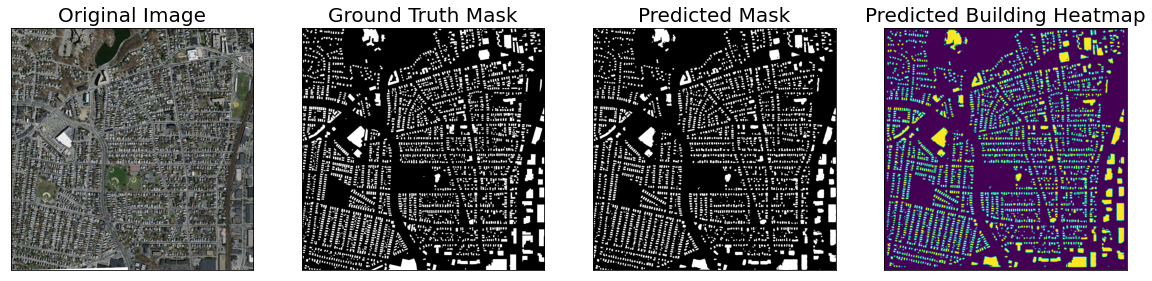}}
		\label{fig:gt2}

	\caption{Example of Predicted Image Vs Original Ground-Truth of U-Net++(efficientNetb4)}

\end{figure}
\begin{table*}[htbp]
	\centering
	\renewcommand{\arraystretch}{1}
	\setlength{\tabcolsep}{15pt}
	\caption{Performance Comparison with Existing Work \label{tab:comp}}
	\resizebox{\textwidth}{!}{
		\begin{tabular}{cllclll}
			\toprule
			\textbf{Year} & \textbf{Reference} & \textbf{Dataset} & \textbf{Network(Type)}  & \textbf{Matrices}\\
                \midrule
				2021 & Abdollahi~\etal~\cite{abdollahi2021integrating} &  AIRS & MultiRes-UNet+ with edge information & \makecell[l]{MCC-95.73
				Precision:$85.8$\%\\ F1: 96.98\% \\ \ IoU: - 94.13\% } \\
			\midrule
				2021 & LEILEI XU~\etal~\cite{xu2021ha} &  challenge dataset & Urban 3D & \makecell[l]{Accuracy: $97.7$\%
				Kappa-80.21\%\\ F1-81.24\%\\ IoU: 70.66\%} \\
			\midrule
				2021 & LEILEI XU~\etal~\cite{xu2021ha} &  WHU dateset & Urban 3D &  \makecell[l]{IoU:72.74\%\\Kappa:79.42\%\\Ins F1:79.32\%} \\
			\midrule
				2021 & Li~\etal~\cite{li2021attention} &   5 WorldView-2 satellite remote sensing image datasets  & attention-enhanced U-Net network & \makecell[l]{Accuracy-96.96\%\\F1 Score-81.47\%\\Recal-82.72\%\\IoU-68.72\%} \\
			\midrule
				2020 & Delibasoglu~\etal~\cite{delibasoglu2020improved} &   Ikonos and Quickbird pan-sharpened satellite images dataset &  Inception UNet-v2 & \makecell[l]{Precision-88.97\%\\F1-82.03\%\\Recall-83.78\%\\Kappa-80.28\%} \\
			\midrule
            
				2020 & Delibasoglu~\etal~\cite{delibasoglu2020improved} &   Massachusetts building dataset &  Inception UNet-v2 & \makecell[l]{Precision- 73.69\%\\F1-78.39\%\\Recall-75.68\%\\Kappa-71.14\%} \\
			\midrule
            
				2020 & Khoshboresh~\etal~\cite{khoshboresh2020multiscale} &   Indiana &  deep dilated CNN & \makecell[l]{Fi score- 83\% \\IoU-72\%} \\
			\midrule
   		
				2020 & Khoshboresh~\etal~\cite{khoshboresh2020multiscale} &   WHU-I &  deep dilated CNN & \makecell[l]{Fi score- 80\% \\IoU-67\%} \\
			\midrule
            
				2019 & Yi~\etal~\cite{yi2019semantic} &  Aerial images with a spatial resolution of 0.075m are collected from the public source &  DeepResUnet & \makecell[l]{Precision-94.01\%\\Recall- 93.28\%\\F1- 93.64 \%\\Kappa-91.76\%\\OA-97.09\%} \\
			\midrule
   		
				2020 & Wagner~\etal~\cite{wagner2020u} &  WorldView-3 images &  U-net-id & \makecell[l]{accuracy-97.67\%\\Precision-0.936\%\\Recall-0.939\% \\Fi score-0.937\% \\IoU mean-0.582\%\\IoU median- 0.694\%\\ Detection rate-97.67\%} \\
			\midrule
   		
				2018 & Lu~\etal~\cite{lu2018dual} &  Inria &  dual-resolution U-Net  & \makecell[l]{IoU-72.45\%} \\
			
   		\midrule
				2018 & Lu~\etal~\cite{lu2018dual} &  Massachusetts buildings  & dual-resolution U-Net  & \makecell[l]{IoU-71.03\%} \\
			\midrule
                2022 & Alsabhan~\etal~\cite{alsabhan2022automatic} & Massachusetts building dataset & Unet-ResNet50 & \makecell[l]{dice loss-10.8%
                IoU score-82.2\%\\Accuracy-90.2\%\\F1 score-90.0\%} \\
			\midrule
				2022 & Alsabhan~\etal~\cite{alsabhan2022automatic} &  Massachusetts building dataset & Unet & \makecell[l]{dice loss-3.3277\%\\IoU score-23.16\% \\Accuracy-71.9\%\\F1 score-60.3\%\\} \\
			\midrule
				2022 & Alsabhan~\etal~\cite{alsabhan2022automatic} & Massachusetts building dataset & FCN & \makecell[l]{OA : 93.37\%\\IoU:69.47\%\\F1:81.98\%\\} \\
			\midrule
			
				2022 & Alsabhan~\etal~\cite{alsabhan2022automatic} & Massachusetts building dataset & SegNET & \makecell[l]{OA : 93.84\%\\IoU:72.1\%\\F1:93.78\%\\} \\
			\midrule
			
				2022 & Alsabhan~\etal~\cite{alsabhan2022automatic} & Massachusetts building dataset & DeepLab v3 & \makecell[l]{OA : 93.01\%\\IoU:68.55\%\\F1:81.34\%\\} \\
			\midrule
			
				2022 & Alsabhan~\etal~\cite{alsabhan2022automatic} & Massachusetts building dataset & ENRU-Net without APNB & \makecell[l]{OA : 94.12\%\\IoU:72.77\%\\F1:84.24\%\\} \\
    \midrule
				2022 & Alsabhan~\etal~\cite{alsabhan2022automatic} & Massachusetts building dataset & ENRU-Net & \makecell[l]{OA : 94.18\%\\IoU:73.02\%\\F1:84.41\%\\} \\
			\midrule
            \textbf{2022} & \textbf{Ours}  &  \textbf{Massachusetts building dataset} &  \textbf{UNET++ (efficientNetb0)} & \makecell[l]{\bf Mean F1:$58.25$\%\\ \bf Mean IoU:$81.63$\%\\ \bf Mean Precision:$87.18$\%\\ \bf Mean Accuracy :$90.8998$\%\\ \bf Mean Recall:$92.62$\%\\ } 
				\\
    \midrule
            \textbf{2022} & \textbf{Ours}  &  \textbf{Massachusetts building dataset} &  \textbf{UNET++ (efficientNetb1)} & \makecell[l]{\bf Mean F1:$60.34$\%\\ \bf Mean IoU:$82.43$\%\\ \bf Mean Precision:$91.0$\%\\ \bf Mean Accuracy :$90.95$\%\\ \bf Mean Recall:$93.01$\%\\ } 
				\\
    \midrule
            \textbf{2022} & \textbf{Ours}  &  \textbf{Massachusetts building dataset} &  \textbf{UNET++ (efficientNetb2)} & \makecell[l]{\bf Mean F1:$63.0$\%\\ \bf Mean IoU:$82.83$\% \\ \bf Mean Precision:$91.6$\%\\ \bf Mean Accuracy :$90.97$\%\\ \bf Mean Recall:$94.0$\%\\ }
				\\
    \midrule
            \textbf{2022} & \textbf{Ours}  &  \textbf{Massachusetts building dataset} &  \textbf{UNET++ (efficientNetb3)} & \makecell[l]{\bf Mean F1:$64.65$\%\\ \bf Mean IoU:$83.12$\%\\ \bf Mean Precision:$93.0$\%\\ \bf Mean Accuracy :$91.0$\%\\ \bf Mean Recall:$94.46$\%\\ } 
				\\
    \midrule
            \textbf{2022} & \textbf{Ours}  &  \textbf{Massachusetts building dataset} &  \textbf{UNET++ (efficientNetb4)} & \makecell[l]{\bf Mean F1:$68.0$\%\\ \bf Mean IoU:$88.32$\%\\ \bf Mean Precision:$93.2$\%\\ \bf Mean Accuracy :$92.23$\%\\ \bf Mean Recall:$94.6$\%\\ } 
				\\
   
			\bottomrule
		\end{tabular}}
  \label{performance_table}
\end{table*}

\section{Conclusion}
Performance investigation of a total of $5$ efficientNet based U-Net++ models for extracting building from remote sensing images has been carried out in this paper. For improved accuracy and efficiency, the proposed design combines use of deep supervision, densely connected redesigned skip connections of U-Net++, and the compound scaling technique algorithm of efficientNet. Experimental results showed that efficientNetb4 based U-Net++ had the best performance among all the variant by achieving mean accuracy, iou and precision of 
$92.23$\%, $88.32$\% and $93.2$\% respectively.
Even though the segmentation result produced by our suggested method was excellent, there are still some issues, such as a poor identification effect of nearby buildings, mistaking shadows for structures, and an inability to recognize buildings that are covered in vegetation. Additionally, there is potential for advancement in terms of the precision of the training dataset and validation dataset by increasing their volume and integrating self-supervised attention mechanism in the further study.

\bibliographystyle{style}
\bibliography{reference}

\begin{thebibliography}{10}
\providecommand{\url}[1]{#1}
\csname url@samestyle\endcsname
\providecommand{\newblock}{\relax}
\providecommand{\bibinfo}[2]{#2}
\providecommand{\BIBentrySTDinterwordspacing}{\spaceskip=0pt\relax}
\providecommand{\BIBentryALTinterwordstretchfactor}{4}
\providecommand{\BIBentryALTinterwordspacing}{\spaceskip=\fontdimen2\font plus
\BIBentryALTinterwordstretchfactor\fontdimen3\font minus
  \fontdimen4\font\relax}
\providecommand{\BIBforeignlanguage}[2]{{%
\expandafter\ifx\csname l@#1\endcsname\relax
\typeout{** WARNING: IEEEtran.bst: No hyphenation pattern has been}%
\typeout{** loaded for the language `#1'. Using the pattern for}%
\typeout{** the default language instead.}%
\else
\language=\csname l@#1\endcsname
\fi
#2}}
\providecommand{\BIBdecl}{\relax}
\BIBdecl

\bibitem{abdollahi2021integrating}
A.~Abdollahi and B.~Pradhan, ``Integrating semantic edges and segmentation
  information for building extraction from aerial images using unet,''
  \emph{Machine Learning with Applications}, vol.~6, p. 100194, 2021.

\bibitem{xu2021ha}
L.~Xu, Y.~Liu, P.~Yang, H.~Chen, H.~Zhang, D.~Wang, and X.~Zhang, ``Ha u-net:
  Improved model for building extraction from high resolution remote sensing
  imagery,'' \emph{IEEE Access}, vol.~9, pp. 101\,972--101\,984, 2021.

\bibitem{alsabhan2022automatic}
W.~Alsabhan and T.~Alotaiby, ``Automatic building extraction on satellite
  images using unet and resnet50,'' \emph{Computational Intelligence and
  Neuroscience}, vol. 2022, 2022.

\bibitem{li2021attention}
C.~Li, L.~Fu, Q.~Zhu, J.~Zhu, Z.~Fang, Y.~Xie, Y.~Guo, and Y.~Gong, ``Attention
  enhanced u-net for building extraction from farmland based on google and
  worldview-2 remote sensing images,'' \emph{Remote Sensing}, vol.~13, no.~21,
  p. 4411, 2021.

\bibitem{delibasoglu2020improved}
I.~Delibasoglu and M.~Cetin, ``Improved u-nets with inception blocks for
  building detection,'' \emph{Journal of Applied Remote Sensing}, vol.~14,
  no.~4, pp. 044\,512--044\,512, 2020.

\bibitem{khoshboresh2020multiscale}
M.~Khoshboresh-Masouleh, F.~Alidoost, and H.~Arefi, ``Multiscale building
  segmentation based on deep learning for remote sensing rgb images from
  different sensors,'' \emph{Journal of Applied Remote Sensing}, vol.~14,
  no.~3, pp. 034\,503--034\,503, 2020.

\bibitem{yi2019semantic}
Y.~Yi, Z.~Zhang, W.~Zhang, C.~Zhang, W.~Li, and T.~Zhao, ``Semantic
  segmentation of urban buildings from vhr remote sensing imagery using a deep
  convolutional neural network,'' \emph{Remote sensing}, vol.~11, no.~15, p.
  1774, 2019.

\bibitem{wagner2020u}
F.~H. Wagner, R.~Dalagnol, Y.~Tarabalka, T.~Y. Segantine, R.~Thom{\'e}, and
  M.~C. Hirye, ``U-net-id, an instance segmentation model for building
  extraction from satellite images—case study in the joan{\'o}polis city,
  brazil,'' \emph{Remote Sensing}, vol.~12, no.~10, p. 1544, 2020.

\bibitem{MnihThesis}
V.~Mnih, ``Machine learning for aerial image labeling,'' Ph.D. dissertation,
  University of Toronto, 2013.

\bibitem{zhou2018unet}
Z.~Zhou, M.~M.~R. Siddiquee, N.~Tajbakhsh, and J.~Liang, ``Unet++: A nested
  u-net architecture for medical image segmentation,'' 2018.

\bibitem{lu2018dual}
K.~Lu, Y.~Sun, and S.-H. Ong, ``Dual-resolution u-net: Building extraction from
  aerial images,'' in \emph{2018 24th International Conference on Pattern
  Recognition (ICPR)}.\hskip 1em plus 0.5em minus 0.4em\relax IEEE, 2018, pp.
  489--494.

\end{thebibliography}
\end{document}